\documentclass[runningheads]{llncs}

\usepackage{eccv}
\usepackage{eccvabbrv}

\usepackage{graphicx}
\usepackage{multirow}
\usepackage{multicol}
\usepackage{booktabs}
\usepackage[export]{adjustbox}
\usepackage{color}
\usepackage{colortbl}

\usepackage[accsupp]{axessibility}
\usepackage[breaklinks,colorlinks,citecolor=eccvblue]{hyperref}
\usepackage{orcidlink}
\usepackage{subcaption}

\begin{document}

\title{Transfer Learning from Simulated to Real Scenes for Monocular 3D Object Detection}
\titlerunning{Transfer Learning for Monocular 3D Object Detection}

\author{
Sondos Mohamed\thanks{Equal contribution}\inst{1}\orcidlink{0009-0002-5647-6895} \and
Walter Zimmer*\inst{2}\orcidlink{0000-0003-4565-1272} \and
Ross Greer\inst{3}\orcidlink{0000-0001-8595-0379} \and
Ahmed Alaaeldin Ghita\inst{2}\orcidlink{0000-0003-3702-8042} \and
Modesto Castrillón-Santana\inst{4}\orcidlink{0000-0002-8673-2725} \and
Mohan Trivedi\inst{5}\orcidlink{0000-0002-0937-6771}  \and
Alois Knoll\inst{2}\orcidlink{0000-0003-4840-076X}  \and
Salvatore Mario Carta\inst{1}\orcidlink{0000-0001-9481-511X}\and
Mirko Marras\inst{1}\orcidlink{0000-0003-1989-6057} 
}

\authorrunning{S.~Mohamed et al.}

\institute{
University of Cagliari, Cagliari, Italy \\ \email{\{sondos.mohamed,salvatore,mirko.marras\}@unica.it} \and
Technical University of Munich, Munich, Germany  \\ \email{\{walter.zimmer,ahmed.ghita,k\}@tum.de}
\and
University of California, Merced, USA  \\ \email{rossgreer@ucmerced.edu}
\and
Universidad de Las Palmas de Gran Canaria, Las Palmas de Gran Canaria, Spain \\ \email{modesto.castrillon@ulpgc.es} 
\and
University of California, San Diego, USA  \\ \email{mtrivedi@ucsd.edu}
}

\maketitle

\begin{abstract}
Accurately detecting 3D objects from monocular images in dynamic roadside scenarios remains a challenging problem due to varying camera perspectives and unpredictable scene conditions. This paper introduces a two-stage training strategy to address these challenges. Our approach initially trains a model on the large-scale synthetic dataset, \textit{RoadSense3D}, which offers a diverse range of scenarios for robust feature learning. Subsequently, we fine-tune the model on a combination of real-world datasets to enhance its adaptability to practical conditions. Experimental results of the \textit{Cube R-CNN} model on challenging public benchmarks show a remarkable improvement in detection performance, with a mean average precision rising from 0.26 to 12.76 on the \textit{TUM Traffic A9 Highway} dataset and from 2.09 to 6.60 on the \textit{DAIR-V2X-I} dataset, when performing transfer learning. Code, data, and qualitative video results are available on the project website: \url{https://roadsense3d.github.io}.
\end{abstract}

\section{Introduction}
\label{sec:intro}
The development of smart cities has become increasingly crucial as urban areas expand and face complex challenges in traffic management and safety. Intersections, for example, are responsible for 40\% of major injuries in Canada and vehicle crashes in the United States \cite{DBLP:conf/itsc/CarrilloW21}. Integrating advanced technologies, such as cameras, into monitoring systems is central to the smart city concept. In China alone, around 200 million outdoor cameras are deployed in their Skynet project \cite {DBLP:journals/tis/SeoaneV24}. While LiDAR and radar are also utilized for traffic monitoring \cite{zimmer2022realdomain,zimmer2023real}, cameras present a cost-effective solution with an extensive range of perception, making them more widely deployable and affordable. 

Accurate detection of objects within monocular camera images is paramount for facilitating intelligent monitoring and effective decision-making \cite{DBLP:conf/percom/AtzoriBCFP21}. Recent advancements in deep learning have fueled a growing interest in 2D/3D object detection approaches. Traditional one-step and two-step 2D object detection methods which predominantly analyze pixel-level information \cite{DBLP:conf/nips/RenHGS15,DBLP:conf/cvpr/CaoCAKP020,DBLP:journals/corr/HeGDG17,DBLP:conf/iccv/LiCWZ19,DBLP:conf/cvpr/RedmonDGF16,DBLP:conf/eccv/LiuAESRFB16,DBLP:conf/eccv/LawD18,DBLP:journals/corr/abs-1904-07850,DBLP:conf/iccv/LinGGHD17}, and, more recently, anchor-free and transformer methods, have been involved in traffic monitoring and applications \cite{ greer2023salient, greer2023robust, greer2024patterns, greer2023pedestrian}. However, methods that provide only 2D detections are limited in providing precise real-world distance measurements between objects as well as the egocentric object. This limitation underscores the necessity for a more comprehensive understanding of the scene and the development of advanced 3D object detection capabilities. Recently, there has been growing interest in infrastructure 3D datasets \cite{cress2022a9,zimmer2023tumtraf,zimmer2024tumtrafv2x,zimmer2024tumtrafv2xsupplement,zhou2024warm3d,DBLP:conf/cvpr/YeSLSLW0D22}.   

Recent monocular 3D models have shown impressive results. However, generalization remains a significant challenge, and most models are domain-specific \cite{DBLP:conf/um/CartaSMMPSSZ24 ,DBLP:journals/corr/abs-2310-05366 }. Exposing these models to a diverse spectrum of datasets with varying factors can enhance their robustness. Models like \textit{Cube R-CNN}\cite{DBLP:conf/cvpr/Brazil0SR0G23} and \textit{UniMode}\cite{li2024unimode}, trained on a wide range of indoor and outdoor datasets, exemplify this approach. Despite their success, these models still face challenges in unfamiliar environments, e.g., roadside scenarios \cite{DBLP:conf/um/CartaSMMPSSZ24}. Another work, \textit{MonoUNI} \cite{jinrang2023monouni}, integrates vehicle and infrastructure data, which adds long-range perception capability. It is evaluated on five benchmarks: \textit{Rope3D}\cite{DBLP:conf/cvpr/YeSLSLW0D22}, \textit{DAIR-V2X-I}, \textit{KITTI} \cite{DBLP:conf/cvpr/GeigerLU12}, \textit{Waymo} \cite{sun2020scalability}, and \textit{nuScenes} \cite{caesar2020nuscenes}. However, separate training for the vehicle and infrastructure domains is still necessary, and a hybrid training that combines both domains is not yet possible. Also, despite the differences between \textit{DAIR-V2X-I}\cite{DBLP:journals/corr/abs-2204-05575} and \textit{Rope3D}\cite{carta2024roadsense3d} as roadside datasets, they share similarities in their view. On the other hand, the model requires calibration information during the inference, which is often lacking in roadside infrastructure cameras. Consequently, there is a demand for monocular 3D models with zero-shot capability to produce an object's 3D position, size, and orientation (9 attributes per object). 

While these models show high accuracy under typical (driving) conditions, their performance degrades significantly when encountering roadside scenarios, such as vehicles that are tilted or overturned due to an accident, primarily due to the limitations in the data annotation process. Specifically, most autonomous driving models predominantly rely on the yaw angle \cite{greer2021trajectory}, often neglecting the roll and pitch angles because they are zeros. However, these angles are crucial in accurately detecting objects at slight elevations, such as in roadside scenarios. To address limitations, in this paper, we conduct comprehensive transfer learning experiments using the \textit{Cube R-CNN} model, transitioning from synthetic datasets such as \textit{RoadSense3D} \cite{DBLP:conf/um/CartaSMMPSSZ24} to real-world datasets like \textit{TUM Traffic A9 Highway} (\textit{TUMTraf-A9}) \cite{cress2022a9} and \textit{DAIR-V2X-I} \cite{DBLP:journals/corr/abs-2204-05575}. In these experiments, we incorporate pitch and roll into both the training and testing phases. The real-world datasets are collected from multiple cities, each with distinct infrastructure configurations, ensuring that the model is exposed to a wide range of urban environments for improved generalization. Through extensive evaluation across these three real-world datasets, we demonstrate that transfer learning improves the $3D~mAP$ results from 0.26 to 12.76 on the \textit{TUMTraf-A9} dataset and from 2.09 to 6.60 on the \textit{DAIR-V2X-I} dataset, when transitioning from simulated to real scenes. We provide model code, datasets, and qualitative video results on our project website: \url{https://roadsense3d.github.io}.
    
\section{Related Work} 
Our work builds upon prior research focusing on data collection using monocular cameras, as well as methods for detecting 3D bounding boxes from these images.

\subsection{Datasets for 3D Monocular Object Detection}
Concerted efforts have been made to collect datasets that include 3D bounding box annotations (Table \ref{tab:datasets}). The main factors characterizing these datasets are the domain (vehicle, roadside, or other infrastructure positions), the type (real or synthetic data), and the image characteristics and quantity.

Several datasets are central to the vehicle view domain \cite {DBLP:conf/cvpr/GeigerLU12,DBLP:conf/cvpr/HuangCGCZWLY18,caesar2020nuscenes,DBLP:conf/cvpr/ChangLSSBHW0LRH19,DBLP:conf/icra/PatilMGC19,astar-3d,sun2020scalability,DBLP:conf/nips/MaoNJLCLLY0LYXX21}. Among them, \textit{KITTI}\cite{DBLP:conf/cvpr/GeigerLU12} is a pioneering and widely utilized dataset that provides benchmarks for various vision-related tasks, including 3D object detection and localization. Datasets such as \textit{nuScenes} \cite{caesar2020nuscenes}, \textit{Argoverse} \cite{DBLP:conf/cvpr/ChangLSSBHW0LRH19}, and \textit{Waymo Open} \cite{sun2020scalability} have covered around 1,000 driving scenes, with the latter distinguishing itself by offering the largest number of 3D bounding boxes. \textit{ONCE} provides the largest number of frames. \textit{Argoverse} \cite{DBLP:conf/cvpr/ChangLSSBHW0LRH19} is notable for including HD maps. Significant advancements in 360-degree camera viewpoints are demonstrated by datasets like \textit{H3D} \cite{DBLP:conf/icra/PatilMGC19}, \textit{nuScenes} \cite{caesar2020nuscenes}, \textit{ONCE} \cite{DBLP:conf/nips/MaoNJLCLLY0LYXX21}, and \textit{Argoverse} \cite{DBLP:conf/cvpr/ChangLSSBHW0LRH19}. The \textit{A*3D} \cite{astar-3d} focuses on the high object density and heavy occlusions. Additionally, \textit{ApolloScape} provides image-based 3D instance segmentation and includes object tracking. Despite the multimodality and diverse tasks offered by these datasets, they still focus on cameras mounted on the car, making them more susceptible to obstacles and short-term events. Therefore, other datasets are targeting long-term prediction. \textit{Ko-PER} \cite{DBLP:conf/itsc/StrigelMSWD14} is one of the pioneering infrastructure datasets, providing 3D object detection data from a permanent intersection monitoring system. Datasets such as \textit{TUMTraf Intersection} \cite{cress2022a9}, \textit{DAIR-V2X-I} \cite{DBLP:journals/corr/abs-2204-05575}, \textit{V2X-Seq} \cite{yu2023v2x}, and \textit{Rope3D}\cite{DBLP:conf/cvpr/YeSLSLW0D22} address various intersection scenarios. \textit{DAIR-V2X-I}, \textit{V2X-Seq}, and recently, \textit{TUMTraf V2X} \cite{zimmer2024tumtrafv2x,zimmer2024tumtrafv2xsupplement} provide views from the vehicle and the roadside infrastructure. \textit{V2X-Seq} offers data for sequential perception  which is derived from DAIR-V2X and trajectory forecasting, while \textit{TUMTraf-A9}~\cite{cress2022a9} covers unsequential highway scenarios. Furthermore, BoxCars \cite{Sochor2018} provides a large-scale intersection dataset as 2D object projections of 3D boxes.

Besides real data, there are several simulated roadside datasets. \textit{TUMTraf Synthetic} \cite{zhou2024warm3d} is built using the \textit{CARLA} simulator \cite{DBLP:conf/corl/DosovitskiyRCLK17} and follows the \textit{KITTI} format\cite{DBLP:conf/cvpr/GeigerLU12}. The camera position and orientation are automatically varied for each frame. It includes diverse weather conditions and provides annotations for ten object classes. The dataset also includes ground truth data for semantic segmentation, depth maps, and RGB images. On the other hand, 
\textit{RoadSense3D} \cite{carta2024roadsense3d} is composed of 35 intersection areas collected from 7 \textit{CARLA} towns. It is considered the largest roadside dataset in terms of the number of images and 3D bounding boxes, featuring more than 9M 3D bounding boxes and approximately 1.4M frames and considering pitch angles. 

The literature shows that real-world datasets including a roadside view may not be large enough to train solid models, while synthetic roadside datasets, although larger, introduce a domain gap when applied to real-world scenarios. Our intuition in this work is to leverage synthetic roadside datasets for initial training, to learn foundational features, and then fine-tune the model on smaller real-world roadside datasets for domain-specific accuracy.

\begin{table}[t]
\caption{Comparison of existing publicly available datasets for the vehicle and roadside infrastructure domains, including their year of release, domain, type, labeling range, RGB resolution, number of RGB images, number of 3D boxes, presence of rain/night data, and availability to the public. 
}
\label{tab:datasets}
\vspace{-6mm}
\centering
\begin{center}
\resizebox{1.\linewidth}{!}{
\begin{tabular}{lrrrrrrrrr}
\toprule
\textbf{Dataset}     & \textbf{Year} & \textbf{Domain}     & \textbf{Type}    & \textbf{Range} & \textbf{Resolution} & \textbf{Images} & \textbf{3D Boxes} & \begin{tabular}[c]{@{}l@{}}\textbf{Rain}/\\ \textbf{Night}\end{tabular} & \textbf{Public} \\
\midrule
KITTI \cite{DBLP:conf/cvpr/GeigerLU12}       & 2013 & Vehicle       & Real      & 70m   & 1392x512   & 15K    & 80K      & No/No                                                 & Yes     \\ 
KoPER \cite{DBLP:conf/itsc/StrigelMSWD14}     & 2014 & Roadside & Real      & -     & 656x494    & -      & -        & No/No                                                 & Yes     \\ 
Apolloscape \cite{DBLP:conf/cvpr/HuangCGCZWLY18} & 2018 & Vehicle       & Real      & \underline{420m}  & \textbf{3384x2710}  & 144K   & 70K      & No/Yes                                                & Yes     \\ 
BoxCars \cite{Sochor2018}    & 2018 & Roadside & Real      & -     & 128x128    & 116K   & 116K     & No/No                                                 & Yes     \\ 
nuScenes \cite{caesar2020nuscenes}    & 2019 & Vehicle       & Real      & 75m   & 1600x900   & \underline{1.4M}   & 1.4M     & Yes/Yes                                               & Yes     \\ 
Argoverse \cite{DBLP:conf/cvpr/ChangLSSBHW0LRH19}   & 2019 & Vehicle       & Real      & 200m  & 1920x1200  & 22K    & 993K     & Yes/Yes                                               & Yes     \\ 
H3D \cite{DBLP:conf/icra/PatilMGC19}        & 2019 & Vehicle       & Real      & 100m  & 1920x1200  & 27.7K  & 1M       & No/No                                                 & Yes     \\ 
A*3D  \cite{astar-3d}      & 2020 & Vehicle       & Real      & 100m  & \underline{2048x1536}  & 39K    & 230K     & Yes/Yes                                               & Yes     \\ 
Waymo Open \cite{sun2020scalability}  & 2020 & Vehicle       & Real      & 75m   & 1920x1080  & 230K   & \textbf{12M}      & Yes/Yes                                               & Yes     \\ 
DAIR-V2X-I \cite{DBLP:journals/corr/abs-2204-05575}    & 2021 & Vehicle/Other & Real      & 200m  & 1920x1080  & 71K    & 1.2M     & -/Yes                                                 & Yes      \\ 
BAAI-VANJEE \cite{DBLP:journals/corr/abs-2105-14370} & 2021 & Roadside & Real      & -     & 1920x1080  & 5K     & 74K      & Yes/Yes                                               & No      \\ 
ONCE \cite{DBLP:conf/nips/MaoNJLCLLY0LYXX21}      & 2021 & Vehicle       & Real      & 200m  & 1920x1080  & \textbf{7M}     & 417K     & Yes/Yes                                               & Yes     \\ 
Rope3D \cite{DBLP:conf/cvpr/YeSLSLW0D22}      & 2022 & Roadside & Real      & 200m  & 1920x1200  & 50K    & 1.5M     & Yes/Yes                                               & No      \\ 
TUMTraf-A9~\cite{cress2022a9}      & 2022 & Roadside & Real      & \textbf{700m}  & 1920x1200  & 1k    & 15k     & Yes/Yes                                               & Yes      \\ 
RoadSense3D~\cite{yu2023v2x}      & 2023 & Roadside  & Synthetic      & 200 & 1920x1080  &  \underline{1.4M}   & \underline{9M}  & Yes/Yes                                               & Yes      \\ 
V2X-Seq~\cite{yu2023v2x}      & 2023 & Vehicle/Other  & Real      & 200 & 1920x1080  & 15k    & 10.45k  & Yes/Yes                                               & Yes      \\ 
TUMTraf Intersection~\cite{cress2022a9}      & 2023 & Roadside & Real      & 120m  & 1920x1200  & 4.8k    & 62.4k     & Yes/Yes                                               & Yes      \\ 
TUMTraf V2X~\cite{zimmer2024tumtrafv2x,zimmer2024tumtrafv2xsupplement}      & 2024 & Vehicle/Roadside & Real      & 200m  & 1920x1200  & 5k    & 30k     & Yes/Yes                                               & Yes      \\ 
TUMTraf Vehicle~\cite{zimmer2024tumtrafv2x,zimmer2024tumtrafv2xsupplement}      & 2024 & Vehicle & Real      & 200m  & 1920x1200  & 1k    & 30k     & Yes/Yes                                               & Yes      \\ 
TUMTraf Synthetic~\cite{zhou2024warm3d}      & 2024 & Roadside & Synthetic      & 200m  & 1920x1200  & 24k    & 240k     & Yes/Yes                                               & Yes      \\ 
\bottomrule
\end{tabular}}
\end{center}
\end{table}

\subsection{Methods for Monocular 3D Object Detection}
Given the abundance of datasets, several methods have been proposed in the literature to address the 3D bounding box detection task (Table \ref{tab:methods}). Such methods can be categorized into two main classes: those utilizing 2D features, referred to as result lifting methods, and those leveraging 3D features, which encompass both feature lifting and data lifting methods. \cite{DBLP:journals/pami/MaOSR24}. 

Data lifting methods convert the entire 2D image into a 3D representation. A notable example is the pseudo-LiDAR method, which generates point cloud data from images before applying a LiDAR-based model for detection \cite{DBLP:conf/iclr/YouWCGPHCW20}. Although data lifting methods yield promising results, they are computationally intensive. Consequently, methods with lower computational requirements have been developed, which fall into two categories: result lifting and feature lifting. Result lifting methods transform 2D detections into 3D by estimating depth to recover the corresponding 3D location from image points \cite{DBLP:conf/cvpr/Brazil0SR0G23}. Feature lifting methods extract 2D features, lift them into 3D space, and then predict 3D objects by transforming image features into 3D voxel grids or orthographic features \cite{DBLP:conf/bmvc/RoddickKC19}. DETR3D \cite{wang2022detr3d}, MonoDLE \cite{DBLP:conf/cvpr/MaZ0ZYLO21}, GUPNet \cite{DBLP:conf/iccv/LuMYZL0YO21}, MonoUNI \cite{jinrang2023monouni}, and Cube R-CNN \cite{DBLP:conf/cvpr/Brazil0SR0G23} are examples of result lifting methods. These methods estimate depth from 2D features to recover the corresponding 3D locations. On the other hand, ImVoxelNet \cite{rukhovich2022imvoxelnet} and UniMODE \cite{li2024unimode} are classified as feature lifting methods, where 2D features are lifted into 3D space for object prediction.

Monocular 3D object detection methods can be further categorized by their application context. Methods designed for vehicle view applications operate in dynamic environments, managing varying speeds and rapidly changing perspectives, which necessitates rapid adaptation for accurate depth estimation \cite{wang2022detr3d,DBLP:conf/iccv/LuMYZL0YO21,rukhovich2022imvoxelnet,DBLP:conf/iccv/ShiYCCCK21,DBLP:conf/cvpr/MaZ0ZYLO21,zhou2021monoef,zhang2021objects,DBLP:conf/iccvw/WangZPL21,wang2022probabilistic,DBLP:journals/corr/abs-2207-07933,kumar2022deviant,DBLP:conf/aaai/Liu0022,li2022diversity,jinrang2023monouni,DBLP:conf/aaai/JiangLLWJWHZ24,liu2024ray,li2024unimode}. Roadside view models, in contrast, are deployed in static environments and optimized for long-range detection, typically utilizing elevated cameras with wide fields of view \cite{zimmer2023infradet3d,yang2023bevheight}. Indoor models must handle shorter focal lengths and significant pitch and yaw variations due to the complex orientations of objects \cite{li2024unimode,DBLP:conf/cvpr/Brazil0SR0G23}.

Despite their application context, several methods can be readily adapted to meet the requirements of roadside infrastructure views, the focus of our study in this paper, as they can extract features over broad ranges and handle different camera configurations.. Specifically, MonoUNI \cite{jinrang2023monouni}  addresses challenges in both roadside infrastructure and vehicle domains by introducing normalized depth to mitigate pitch angle and focal length variations. \textit{ImVoxelNet} \cite{rukhovich2022imvoxelnet} generates voxel representations and employs distinct heads for indoor and outdoor environments, accommodating both multi-view and monocular inputs. \textit{Cube R-CNN} \cite{DBLP:conf/cvpr/Brazil0SR0G23} extends \textit{Faster R-CNN} with 3D branches, utilizing virtual depth and mesh representations for robust object detection across indoor and outdoor scenes. \textit{UniMODE} \cite{li2024unimode} adopts a two-stage architecture, addressing grid size challenges and computes loss based on dataset-specific metrics to handle label inconsistencies.

\begin{table*}[!t]
\centering
\caption{Monocular 3D object detection methods sorted by publication year and method name. The scenario types include vehicle (V), roadside (R), and indoor (I).}
\label{tab:methods}
\resizebox{1.\linewidth}{!}{
\begin{tabular}{llcccl}
\toprule
\textbf{Method} & \textbf{Venue \& Year} & \multicolumn{3}{c}{\textbf{Type}} & \textbf{Short Description} \\
 &  & \textbf{V} & \textbf{R} & \textbf{I} & \\
\midrule
DETR3D~\cite{wang2022detr3d} & PMLR 2021 & \checkmark & & & Transformer-based, direct 3D bounding box prediction from images. \\
GUPNet~\cite{DBLP:conf/iccv/LuMYZL0YO21}  & ICCV 2021 & \checkmark & & & Geometry uncertainty propagation, geometry-aware. \\
ImVoxelNet~\cite{rukhovich2022imvoxelnet} & WACV 2021 & \checkmark & & \checkmark & Voxel-based, multi-view fusion for improved 3D understanding. \\
MonoDet~\cite{DBLP:conf/iccv/ShiYCCCK21} & ICCV 2021 & \checkmark & & & Single-stage, anchor-based detection with depth-aware module. \\
MonoDLE~\cite{DBLP:conf/cvpr/MaZ0ZYLO21} & CVPR 2021 & \checkmark & & &  Efficient depth estimation, improved localization with light-weight architecture. \\
MonoEF~\cite{zhou2021monoef} & TPAMI 2021 & \checkmark & & & Edge fusion, enhanced depth and boundary prediction. \\
MonoFlex~\cite{zhang2021objects} & CVPR 2021 & \checkmark & & & Anchor-free, flexible regression head, uncertainty-based keypoint estimation. \\
MonoXiver~\cite{DBLP:conf/iccvw/WangZPL21} & ICCVW 2021 & \checkmark & & & Single-frame, uncertainty-guided feature refinement. \\
PGD~\cite{wang2022probabilistic} & PMLR 2021 & \checkmark & & & Probabilistic depth modeling, geometry-aware detection. \\
\hline
CIE~\cite{DBLP:journals/corr/abs-2207-07933} & CoRR 2022 & \checkmark & & & Contextual information extraction, multi-view consistency. \\
DEVIANT~\cite{kumar2022deviant} & ECCV 2022 & \checkmark & & & Adversarial training, domain adaptation for robust detection. \\
MonoCon~\cite{DBLP:conf/aaai/Liu0022} & AAAI 2022 & \checkmark & & & Anchor-free, context enhancement, auxiliary task integration. \\
MonoDDE~\cite{li2022diversity} & CVPR 2022 & \checkmark & & & Diversity-driven ensemble, improved generalization across domains. \\
\hline
BEVHeight~\cite{yang2023bevheight} & CVPR 2023 & & \checkmark & & BEV-based, height-guided feature extraction for roadside detection. \\
Cube R-CNN~\cite{DBLP:conf/cvpr/Brazil0SR0G23} & CVPR 2023 & \checkmark & & \checkmark & 3D bounding box regression, multi-modal fusion for indoor scenes. \\
MonoDet3D~\cite{zimmer2023infradet3d} & IV 2023 & & \checkmark & & 3D object detection for roadside infrastructure sensors. \\
MonoUNI~\cite{jinrang2023monouni} & NeurIPS 2023 & \checkmark & \checkmark & & Unified architecture for vehicle and roadside detection, domain adaptation. \\
\hline
Far3D~\cite{DBLP:conf/aaai/JiangLLWJWHZ24} & AAAI 2024 & \checkmark & & & Far-field detection, enhanced distance-aware features for long-range perception. \\
RayDN~\cite{liu2024ray} & ECCV 2024 & \checkmark & & & Ray-based depth estimation, novel loss function for accurate localization. \\
UniMODE~\cite{li2024unimode} & CVPR 2024 & \checkmark & & \checkmark & Unified model for vehicle and indoor detection, multi-domain training. \\
\bottomrule
\end{tabular}
}
\end{table*}

\section{Methodology}
In this section, we first mathematically formulate the task of 3D object detection from monocular cameras. Next, we introduce the model selection process according to the roadside scenario. We then describe initial training with synthetic data, which involves detailing the dataset and the method for training the model from scratch. We then elaborate on the fine-tuning phase, discussing the selected real-world datasets and the technical aspects of the fine-tuning process.

\subsection{Problem Definition}
Image-based 3D object detection involves determining the position and shape of objects in three-dimensional space based on a two-dimensional image captured by a camera. To address this, we aim to learn a function, parametrized by $\theta$, that maps a 2D RGB image $i \in \mathcal{I}$, where $\mathcal{I} \subset \mathbb{R}^{H \times W \times 3}$ represents the set of images with height $H$, width $W$ and corresponding camera parameters, to a set of 3D object attributes. Specifically, for each image $i$, the model outputs attributes for each detected object $j$: category $c_{i,j}$, 3D position coordinates $(x_{i,j}, y_{i,j}, z_{i,j})$, dimensions $(h_{i,j}, w_{i,j}, l_{i,j})$, and yaw-pitch-roll orientation angles $(\vartheta_{i,j}, \phi_{i,j}, \psi_{i,j})$. This process can be formalized as:

\begin{equation}
    \small
    f_\theta(i) \rightarrow \left\{ \left( c_{i,j}, (x_{i,j}, y_{i,j}, z_{i,j}), (h_{i,j}, w_{i,j}, l_{i,j}), (\vartheta_{i,j}, \phi_{i,j}, \psi_{i,j}) \right) \mid j = 1, \ldots, N_i \right\}
\end{equation}

\noindent where $N_i$ denotes the number of objects detected in the image $i$.

To learn a function for yielding these 3D object attributes, we use a dataset $\mathcal{D}$ ($M = |\mathcal{D}|$) consisting of images with annotated 3D bounding boxes. Each entry in the training dataset includes an image $i$ and its ground truth attributes $y_i = \left\{ \left( c_{i,j}^*, (x_{i,j}^*, y_{i,j}^*, z_{i,j}^*), (h_{i,j}^*, w_{i,j}^*, l_{i,j}^*), (\vartheta_{i,j}^*, \phi_{i,j}^*, \psi_{i,j}^*) \right) \mid j = 1, \ldots, N_i \right\}$, where the asterisks denotes the true values. The training objective is to optimize the function's parameters $\theta$ to accurately predict these attributes. In other words, we aim to minimize the loss of the predictions given true annotations. Formally:

\begin{equation}
    \small
    \hat{\theta} = \arg\min_{\theta} \frac{1}{M} \sum_{i=1}^{M} \mathcal{L}\left( y_i, f_\theta(i) \right)
\end{equation}

The typical loss function $\mathcal{L}$ combines several components to address the different aspects of the prediction task, including classification ($\mathcal{L}_{\text{cls}}$), position ($\mathcal{L}_{\text{pos}}$), dimension ($\mathcal{L}_{\text{dim}}$), and orientation ($\mathcal{L}_{\text{ori}}$).
The training process seeks to find the function's parameters $\theta$ that minimize this composite loss across the dataset $\mathcal{D}$. By doing so, the function can yield accurate categories, positions, dimensions, and orientations from 2D images, thereby solving the target task.

\subsection{Model Selection}\label{sec:model_selection}
To select image-based 3D object detection methods, we conducted an extensive review of top-tier computer vision conferences (e.g., ECCV, CVPR, ICCV) and journals (e.g., IEEE Transactions on Pattern Analysis and Machine Intelligence, T-PAMI) for publications since 2021. Our review focused specifically on models that utilize end-to-end architectures, avoiding those dependent on auxiliary networks for depth extraction. We also prioritized models that have demonstrated strong performance across multiple datasets. Our selection criteria included the models' ability to demonstrate domain adaptability and their applicability across diverse tasks (see Table \ref{tab:methods}). From this rigorous assessment, we identified four models for further consideration: \textit{MonoUNI} \cite{jinrang2023monouni}, \textit{ImVoxelNet} \cite{rukhovich2022imvoxelnet}, \textit{Cube R-CNN} \cite{DBLP:conf/cvpr/Brazil0SR0G23}, and \textit{UniMODE} \cite{li2024unimode}. Among these, \textit{Cube R-CNN} was chosen for our experiments due to its notable reproducibility and unified training pipeline. Unlike \textit{UniMODE}, which was excluded due to limitations in its training pipeline, \textit{Cube R-CNN} integrates multiple camera coordinate systems and is robust in handling six degrees of freedom in object orientations. In contrast to \textit{ImVoxelNet} and \textit{MonoUNI}, which rely on dataset-specific training strategies, \textit{Cube R-CNN} does not compromise effectiveness across datasets. 

From an architectural perspective, \textit{Cube R-CNN} builds upon the Faster R-CNN framework \cite{DBLP:conf/nips/RenHGS15}, an end-to-end region-based object detection approach. Faster R-CNN employs a backbone network, typically a convolutional neural network (CNN), to transform the input image into a higher-dimensional feature representation. A Region Proposal Network (RPN) then generates regions of interest (ROIs) that signify potential object candidates within the image. These ROIs are processed by a 2D box head, which uses the backbone's feature map to classify the object and refine the 2D bounding box predictions. A cube head that computes the 3D parameters, including central point projection, depth, scaled dimensions, and object-centered orientation, is applied for each detected object.

\subsection{Initial Model Creation}

\subsubsection{Synthetic Dataset Selection.}
In our experiments, we used the \textit{RoadSense3D} \cite{carta2024roadsense3d} synthetic dataset comprising over 9 million labeled 3D objects across 1.4 million frames for model training. As detailed in Table \ref{tab:datasets}, this dataset offers a diverse range of scenarios generated from 35 roadside cameras across seven distinct towns in the \textit{CARLA} simulator \cite{DBLP:conf/corl/DosovitskiyRCLK17}. Key parameters include 1920x1080 image resolution, 40,448 frames per position, camera pitch angles ranging from -25° to -45°, a detection range of 150 meters, and a 120° field of view. To simulate realistic conditions, the dataset incorporates variations in weather (sunny, cloudy, foggy) and time of day (day/night).

We selected the \textit{RoadSense3D} dataset since there is a lack of large real-world datasets that provide sufficient data from roadside environments. Indeed, although datasets like \textit{Rope3D} \cite{DBLP:conf/cvpr/YeSLSLW0D22} include millions of real-world images, they are not publicly available. Training models from scratch requires a substantial amount of data, which is not readily accessible from the real world in this domain since labeling is costly and often requires manual labeling effort \cite{zimmer20193dbat}. Given this, among artificial datasets, \textit{RoadSense3D} is the largest synthetic dataset available and offers comprehensive coverage of various factors.

\subsubsection{Training from Scratch.}
\begin{figure}[t]
 \includegraphics[width=1.\textwidth,trim={0 0.25cm 0 0.25cm},clip,frame]{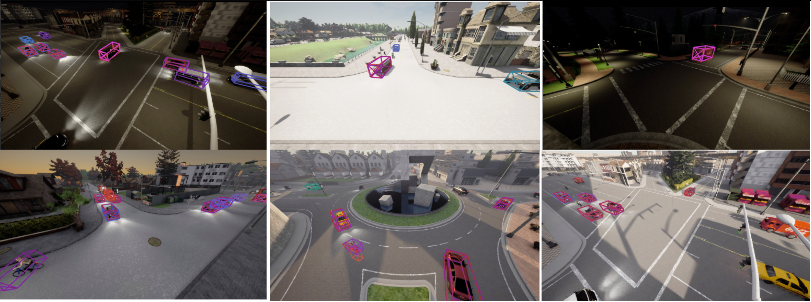}
 \vspace{-5mm}
 \caption{\textbf{Qualitative Results on the Synthetic \textit{RoadSense3D} Test Set.} We show 3D box detections of the \textit{Cube R-CNN} model in the class-specific colors during different lighting and weather conditions.}
 \label{fig:fig_carla}
\end{figure}

We trained \textit{Cube R-CNN} on the \textit{RoadSense3D} synthetic dataset to address object pose variability. The dataset is sequential, with multiple images containing the same objects in different positions. This strategy allowed the model to encounter various objects, locations, and occlusion scenarios. According to the original paper, we split the dataset into training, validation, and testing sets. The model was trained for 250,000 iterations on a single GPU, using a batch size of 4 and a learning rate of 0.0025. We utilized the Stochastic Gradient Descent (SGD) solver, and the model was evaluated every 10,000 iterations. These hyperparameters were chosen to address inconsistencies observed in the original \textit{Cube R-CNN} model's training. Figure \ref{fig:fig_carla} presents qualitative results.

\subsection{Pretrained Model Transfer}

\subsubsection{Real-World Datasets Selection.}
To enable the \textit{Cube R-CNN} model to generalize to real-world scenarios, we fine-tuned it on several diverse datasets that vary in camera setups and environmental conditions. Due to their public availability and size, we selected the \textit{TUMTraf-A9} and \textit{DAIR-V2X-I} datasets. The \textit{TUMTraf-A9} dataset, a subset of the \textit{TUMTraf} dataset family, captures complex highway scenarios in Munich, Germany, under diverse weather and lighting conditions. It features 1,000 labeled frames with 15,000 3D bounding boxes and track IDs and includes data from both LiDAR and multi-view cameras, using 16mm and 50mm focal lengths to monitor traffic across 11 lanes. The \textit{DAIR-V2X-I} dataset, collected in China, represents a large-scale vehicle-infrastructure cooperative autonomous driving dataset, offering over 71,000 LiDAR and camera frames from both infrastructure and vehicle perspectives. 

Pre-processing was needed for the \textit{DAIR-V2X-I} dataset, which provides object annotations in the LiDAR coordinate system. Since \textit{Cube R-CNN} operates in the camera coordinate system, we first transformed the annotations from the LiDAR coordinate system to the camera coordinate system. In the LiDAR system, annotations specify the object's location, dimensions, and yaw rotation across seven degrees of freedom. When projecting these annotations to the camera's view, it is essential to account for the object's rotation relative to the camera. This involves projecting the annotations around all three axes. The transformation process follows the equation $\begin{bmatrix} x & y & w \end{bmatrix} = K \cdot T \cdot \begin{bmatrix} X & Y & Z & 1 \end{bmatrix}$, where the intrinsic matrix $K$ and the extrinsic matrix $T$ are used to project the 3D point $(X, Y, Z)$ in the world coordinate system to the 2D point $(x, y)$ in the camera coordinate system. The variable $w$ serves as a scale factor in this transformation, ensuring the proper projection of points into the camera view. We then utilized a 60/40 ratio for training and testing. For the test, we used sequential perception from \textit{V2X-Seq} \cite{yu2023v2x}. For \textit{TUMTraf-A9}, we applied the same 60/40 ratio for training and testing, using data from both the north and south cameras and incorporating both small and large focal lengths.

\subsubsection{Fine-tuning.}
The model was initially trained on the \textit{RoadSense3D} dataset for 250,000 iterations. Subsequently, we began training with a reduced learning rate of $\alpha=0.0025$, utilizing the DLA34 architecture \cite{DBLP:conf/cvpr/YuWSD18} for feature extraction. Fine-tuning was focused on the head component of the model, which includes both the 2D head and the 3D cube head. The model was further trained for an additional 850,000 iterations.

\section{Experimental Results} 
In this section, we present the results obtained from our transfer learning experiments. First, we examine the impact of transferring from synthetic to real data in a single step, specifically from \textit{RoadSense3D} to \textit{TUMTraf-A9} and \textit{DAIR-V2X-I} separately. Next, we explore the transition from a synthetic dataset to real-world datasets gradually, moving from \textit{RoadSense3D} to \textit{DAIR-V2X-I} and then to \textit{TUMTraf-A9}. The model's performance was assessed on the test set of each real-world dataset using the 3D mean Average Precision ($mAP_{3D}$) under a certain Intersection over Union (IoU) threshold, measuring both detection and localization precision.

\begin{table*}[t]
\centering
\caption{\textbf{Single-Step Dataset Transfer on \textit{TUMTraf-A9}}. We report the 3D mean average precision across the easy, moderate, and hard difficulty levels. Transfer learning involves pre-training on the synthetic \textit{RoadSense3D} dataset and fine-tuning on the real-world \textit{TUMTraf-A9} dataset.}
\label{tab:singlestep-a9}
\resizebox{1.\linewidth}{!}{
\begin{tabular}{l|lll|rrrr}
\toprule
\textbf{Architecture} & \textbf{Pre-Train Set} & \textbf{Fine-Tuning Set} & \textbf{Evaluation Set} & \multicolumn{3}{c}{\bf Difficulty Level} \\
 & & & & \textbf{Easy} & \textbf{Moderate} & \textbf{Hard} \\
\midrule
Cube R-CNN & TUMTraf-A9 Train & - & TUMTraf-A9 Test & 0.26 & 0.26 & 0.26 \\
\rowcolor{gray!10}
Cube R-CNN & RoadSense3D Train & TUMTraf-A9 Train & TUMTraf-A9 Test & \textbf{12.76} & \textbf{12.76} & \textbf{12.76} \\
\bottomrule 
\end{tabular}}
\end{table*}

\begin{table*}[t]
\centering
\caption{\textbf{Single-Step Dataset Transfer on \textit{DAIR-V2X-I}}. We report the 3D mean average precision across the easy, moderate, and hard difficulty levels. Transfer learning involves pre-training on the synthetic \textit{RoadSense3D} dataset and fine-tuning on the real-world \textit{DAIR-V2X-I} dataset.}
\label{tab:singlestep-v2x}
\resizebox{1.\linewidth}{!}{
\begin{tabular}{l|lll|rrrr}
\toprule
\textbf{Architecture} & \textbf{Pre-Train Set} & \textbf{Fine-Tuning Set} & \textbf{Evaluation Set} & \multicolumn{3}{c}{\bf Difficulty Level} \\
 & & & & \textbf{Easy} & \textbf{Moderate} & \textbf{Hard} \\
\midrule
Cube R-CNN & DAIR-V2X-I Train & - & DAIR-V2X-I Test & 2.09   &  2.62  &  2.61  \\
\rowcolor{gray!10}
Cube R-CNN & RoadSense3D Train & DAIR-V2X-I Train & DAIR-V2X-I Test & \textbf{6.60}   & \textbf{8.60} &  \textbf{8.65}   \\
\bottomrule 
\end{tabular}}
\end{table*} 

\begin{figure}[t]
    \centering
    \begin{subfigure}[t]{\textwidth}
        \centering
        \includegraphics[width=\textwidth,trim={0 0.25cm 0 0.25cm},clip,frame]{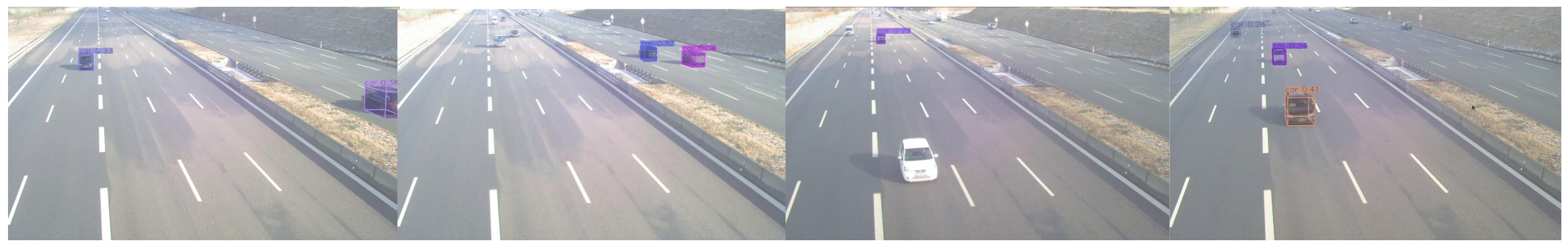}
        \caption{Model trained from scratch on \textit{TUMTraf-A9}.}
        \label{fig:img-a9-only}
    \end{subfigure}
    \vspace{1cm} 
    \begin{subfigure}[t]{\textwidth}
        \centering
        \includegraphics[width=\textwidth,trim={0 0.25cm 0 0.25cm},clip,frame]{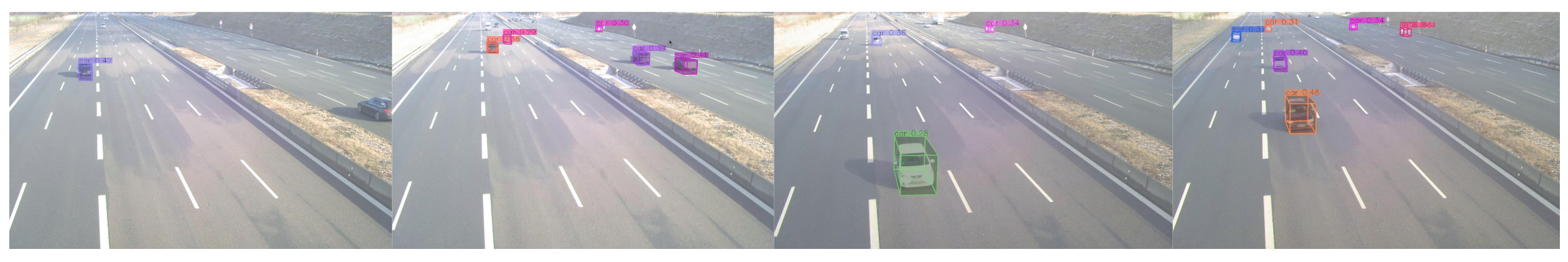}
        \caption{Model pre-trained on \textit{RoadSense3D} and fine-tuned on \textit{TUMTraf-A9}.}
        \label{fig:img-rodsense3d-a9}
    \end{subfigure}
    \vspace{-10mm}
    \caption{\textbf{Qualitative Results of Cube R-CNN on the \textit{TUMTraf-A9} Test Set.} Comparison between the \textit{Cube R-CNN} model trained from scratch on \textit{TUMTraf-A9} (top) and model trained on \textit{RoadSense3D} and fine-tuned on \textit{TUMTraf-A9} (bottom).}
    \label{fig:qual-a9}
\end{figure}


\subsection{Single-Step Dataset Transfer}
In our initial analysis, we investigated the performance of direct transfer learning. To this end, we trained a \textit{Cube R-CNN} model on the full \textit{RoadSense3D} dataset and subsequently fine-tuned it separately on the training sets of \textit{DAIR-V2X-I} and \textit{TUMTraf-A9}, resulting in two fine-tuned models.  To assess the impact of transfer learning, we also trained independent \textit{Cube R-CNN} models from scratch on the training sets of \textit{DAIR-V2X-I} and \textit{TUMTraf-A9}, resulting in two additional models for comparison. We then evaluated the transferability of these models on the test sets of the corresponding real-world dataset, \textit{DAIR-V2X-I} and \textit{TUMTraf-A9}, respectively. Tables \ref{tab:singlestep-a9} and \ref{tab:singlestep-v2x} collect the $mAP_{3D}$ across difficulty levels for the two model instances when evaluated on the test part of the respective real-world dataset, respectively. 

For the \textit{TUMTraf-A9} test set, the model pre-trained on \textit{RoadSense3D} and fine-tuned on \textit{TUMTraf-A9} achieved a $mAP_{3D}$ of 12.76 across easy, moderate, and hard difficulty levels. This represents an increase of 4,808\% compared to the $mAP_{3D}$ of 0.26 for the model trained from scratch on \textit{TUMTraf-A9}. Notably, the values for easy, moderate, and hard difficulty levels are identical (12.76). This occurs because the \textit{TUMTraf-A9} test set lacks occlusions. Without them, which typically increase the detection task complexity, the model does not face additional challenges across difficulty levels. The gains observed quantitatively can be better appreciated in Figure \ref{fig:qual-a9}, which provides qualitative examples of detections for both the considered models.

For the \textit{DAIR-V2X-I} test set, the model fine-tuned with transfer learning achieved $mAP_{3D}$ scores of 6.60 (easy), 8.60 (moderate), and 8.65 (hard), compared to 2.09, 2.62, and 2.61 for the model trained from scratch. These results indicate performance improvements of 215.8\% (easy), 228.6\% (moderate), and 231.4\% (hard), showing the effectiveness of transfer learning across all difficulty levels. The varying gains across difficulty levels also suggest that the benefits of transfer learning are more pronounced in moderate and hard scenarios, likely due to the increased complexity of these cases. 

In summary, these experiments show that transfer learning from synthetic data can substantially improve real-world performance, particularly when the target dataset is small or has complex scenarios.

\begin{table*}[t]
\centering
\caption{\textbf{Multi-Step Dataset Transfer on \textit{TUMTraf-A9}}. We report the 3D mean average precision across the easy, moderate, and hard difficulty levels. Transfer learning involves training on synthetic \textit{RoadSense3D}, then tuning on \textit{DAIR-V2X-I}, and finally on \textit{TUMTraf-A9}.}
\label{tab:multistep-a9}
\resizebox{1.\linewidth}{!}{
\begin{tabular}{l|lll|rrrr}
\toprule
\textbf{Architecture} & \textbf{Pre-Train Set} & \textbf{Fine-Tuning Set} & \textbf{Evaluation Set} & \multicolumn{3}{c}{\bf Difficulty Level} \\
 & & & & \textbf{Easy} & \textbf{Moderate} & \textbf{Hard} \\
\midrule
Cube R-CNN & TUMTraf-A9 Train & - & TUMTraf-A9 Test & 0.26 & 0.26 & 0.26 \\
\rowcolor{gray!10}
Cube R-CNN & RoadSense3D Train &  DAIR-V2X Train $\rightarrow$ TUMTraf-A9 Train & TUMTraf-A9 Test & \textbf{6.26} & \textbf{6.26} &  \textbf{6.26} \\
\bottomrule 
\end{tabular}}
\end{table*}

\subsection{Multi-Step Dataset Transfer}
In our second analysis, we investigated the performance of gradual transfer learning. To this end, we used the \textit{Cube R-CNN} model pre-trained on the full \textit{RoadSense3D} dataset and subsequently fine-tuned it first on the larger training set of \textit{DAIR-V2X-I} and then on the smaller training set of \textit{TUMTraf-A9}. To assess the impact of transfer learning, we also trained independent \textit{Cube R-CNN} models from scratch on the training set of \textit{TUMTraf-A9}, resulting in another model for comparison. We then evaluated the transferability of these models on the test set of \textit{TUMTraf-A9}. Table \ref{tab:multistep-a9} collects the $mAP_{3D}$ across difficulty levels for the two model instances.

Results show that the \textit{Cube R-CNN} model pre-trained on \textit{RoadSense3D} and then fine-tuned sequentially on \textit{DAIR-V2X-I} and \textit{TUMTraf-A9} achieved a 3D mean Average Precision ($mAP_{3D}$) of 6.26 across all difficulty levels on the \textit{TUMTraf-A9} test set. This represents a substantial increase of 2,308\% compared to the 0.26 $mAP_{3D}$ obtained by the model trained from scratch on \textit{TUMTraf-A9}. However, the model fine-tuned directly on \textit{TUMTraf-A9} after pre-training on \textit{RoadSense3D} (Table \ref{tab:singlestep-a9}) achieved a higher $mAP_{3D}$ of 12.76 across all difficulty levels. This indicates that while the multi-step transfer learning approach can improve performance, it still falls short of the results obtained through direct fine-tuning on \textit{TUMTraf-A9}. We conjecture that it is caused by potential domain gaps introduced during the intermediate \textit{DAIR-V2X-I} phase. This intermediate step might cause the model to adapt real-world features from \textit{DAIR-V2X-I} that are less optimal for \textit{TUMTraf-A9}, influencing the quality of the subsequent fine-tuning phase and leading to lower overall performance compared to the more direct fine-tuning approach.

In summary, these experiments show that while multi-step transfer learning offers performance improvements, the direct fine-tuning approach tends to be more effective for maximizing performance on specific real-world datasets.

\section{Conclusion and Future Work}

In this work, we conducted extensive transfer learning experiments using the \textit{Cube R-CNN} model, transitioning from synthetic datasets like \textit{RoadSense3D} to real-world datasets such as \textit{TUMTraf-A9} and \textit{DAIR-V2X-I}. By incorporating pitch and roll into both training and testing phases and evaluating across multiple cities with diverse infrastructure, we demonstrated significant improvements in detection accuracy. Direct transfer learning enhanced the $3D~mAP$ from 0.26 to 12.76 on the \textit{TUMTraf-A9} dataset and from 2.09 to 6.60 on the \textit{DAIR-V2X-I} dataset, showcasing substantial gains in real-world performance. Our findings indicate that while multi-step transfer learning is beneficial, direct fine-tuning on the target dataset yields superior results. This approach bridges the simulation-to-real gap and paves the way for more robust and adaptable models in intelligent transportation systems. The potential applications extend beyond traffic monitoring to include autonomous driving and smart city infrastructure, where accurate and scalable 3D perception is critical for enhancing safety and efficiency.

Future research will explore adapting additional monocular object detection methods to the existing transfer learning framework, incorporating all yaw, pitch, and roll variations to enhance their adaptability to roadside scenarios. We plan to involve detailed inspections to identify scenarios where transfer learning falls short, aiming to inform the development of novel approaches. Furthermore, integrating active learning and knowledge distillation will be pursued to refine the transfer learning process, focusing on selecting only the most informative examples to let the model adapt. Additionally, incorporating these 3D object detection methods into anomaly detection pipelines for real-world smart city applications will be investigated, with particular emphasis on enhancing accident detection and prevention strategies thanks to more precise object detection.

\section*{Acknowledgments}
This research was supported by the Federal Ministry of Education and Research in Germany within the project \textit{AUTOtech.agil}, Grant Number: 01IS22088U. Furthermore, this work has been partially supported by the Autonomous Region of Sardinia through Sardegna Ricerche, within the funding call \textit{Aiuti per Progetti di Ricerca e Sviluppo - Settore ICT (2022)}, under the project \textit{SENTINEL: Sustainable intElligeNce-based soluTion for envIronmental and urbaN survEilLance}, Grant Number: G27H23000140002.

\bibliographystyle{splncs04}
\bibliography{main}
\end{document}